\documentclass[conference]{IEEEtran}
\IEEEoverridecommandlockouts
\usepackage{algorithm}
\usepackage{algorithmic}

\usepackage{xcolor}
\usepackage{amsmath}
\usepackage{booktabs}
\usepackage{multirow}
\usepackage{threeparttable}
\usepackage{subcaption}
\usepackage[section]{placeins}

\usepackage{xurl}
\usepackage[algo2e, ruled,vlined]{algorithm2e}
\usepackage{pbox}
\usepackage{siunitx}

\usepackage{graphicx}
\graphicspath{ {pic/} }
\usepackage{float}
\usepackage{array}
\usepackage{tabularx}
\usepackage{marvosym}
\usepackage{multirow}
\usepackage{array}
\usepackage{enumitem}
\usepackage{booktabs} % For formal tables

\usepackage{graphicx}
\usepackage{textcomp}
\usepackage{xcolor}

\usepackage{pifont}
\usepackage{url}
\usepackage{bm}
\usepackage{makecell}
\usepackage{caption}
\definecolor{lightblue}{RGB}{0, 134, 179}

\makeatletter

\newcommand{\Rmnum}[1]{\expandafter\@slowromancap\romannumeral #1@}
\makeatother

\def\BibTeX{{\rm B\kern-.05em{\sc i\kern-.025em b}\kern-.08em
    T\kern-.1667em\lower.7ex\hbox{E}\kern-.125emX}}

\newtheorem{definition}{Definition}

\begin{document}
\renewcommand{\thefootnote}{\fnsymbol{footnote}}

\title{LISA: Learning-Integrated Space Partitioning Framework for Traffic Accident Forecasting on Heterogeneous Spatiotemporal Data\\
}

\author{
    {Bang An$^{*2}$, Xun Zhou$^{*1}$, Amin Vahedian$^{3}$, Nick Street$^{2}$, Jinping Guan$^{4}$, Jun Luo$^{5}$}\\
}
\maketitle

\footnotetext[1]{Co-first authors with equal contribution.}
\renewcommand{\thefootnote}{\arabic{footnote}}
\footnotetext[1]{Corresponding author, Harbin Institute of Technology, Shenzhen, \texttt{zhouxun2023@hit.edu.cn} This work was done while at the University of Iowa.}
\footnotetext[2]{University of Iowa, \texttt{\{bang-an, nick-street\}@uiowa.edu}}
\footnotetext[3]{Northern Illinois University, \texttt{avahediankhezerlou@niu.edu}}
\footnotetext[4]{Massachusetts Institute of Technology, \texttt{jinpingg@mit.edu}}
\footnotetext[5]{Logistics and Supply Chain MultiTech R\&D Centre, \texttt{jluo@lscm.hk}}

\begin{abstract}
Traffic accident forecasting is an important task for intelligent transportation management and emergency response systems. However, this problem is challenging due to the spatial heterogeneity of the environment. Existing data-driven methods mostly focus on studying homogeneous areas with limited size (e.g. a single urban area such as New York City) and fail to handle the heterogeneous accident patterns over space at different scales. Recent advances (e.g. spatial ensemble) utilize pre-defined space partitions and learn multiple models to improve prediction accuracy. However, external knowledge is required to define proper space partitions before training models and pre-defined partitions may not necessarily reduce the heterogeneity. To address this issue, we propose a novel Learning-Integrated Space Partition Framework (LISA) to simultaneously learn partitions while training models, where the partitioning process and learning process are integrated in a way that partitioning is guided explicitly by prediction accuracy rather than other factors. Experiments using real-world datasets, demonstrate that our work can capture underlying heterogeneous patterns in a self-guided way and substantially improve baseline networks by an average of $13.0\%$.
\end{abstract}


\section{Introduction}

Traffic accidents are major safety concerns in modern society. According to the National Highway Traffic Safety Administration (NHTSA) \cite{2020fatality}, an estimated 42,939 people died in motor vehicle traffic crashes in the United States in 2021, which is the highest number of fatalities since 2005 and the largest annual percentage increase in the history. Given such growth in fatalities, the ability to predict future traffic accidents is critically significant to transportation participants and public safety stakeholders. An accurate and generalizable traffic accident forecasting system can improve transportation management and emergency response deployment. For example, Tennessee Highway Patrol (THP) has used predictive analytics to reduce its average response time by $33\%$ since 2012 and lowered the accident fatality rate by around $8\%$ from 2010 to 2016. \cite{acctool}

However, traffic accident prediction is challenging. First, the factors that result in crashes are complicated, and there exist dependencies between accidents and surrounding attributes such as road conditions and weather conditions. Moreover, such patterns are not likely to be homogeneous through time and space. For example, the factors leading to accidents in the congested downtown area will likely differ from the factors on quiet country roads. Given such complex patterns and heterogeneous environments, it is non-trivial to precisely predict accidents using a single model.

There is a wide range of literature on traffic accident forecasting problems using machine learning methods\cite{HuMingzhi2023SPfR}\cite{BarbaLida2014SSCw}\cite{ChenWeiye2024RfFA}\cite{bergel2013explaining}\cite{CALIENDO2007657}\cite{chang2005data}. However, as classic models have difficulty in learning large datasets and complex functions, they fail to capture complicated factors leading to accidents. More recently, researchers have attempted to use deep learning techniques including recurrent neural networks \cite{Ren2018ADL} and convolutional neural networks \cite{najjar2017combining} to capture temporal changes and spatial correlations. Nevertheless, general-purpose machine learning methods are not designed for predicting traffic accidents, thus performing poorly when inherent dependencies and spatial heterogeneity exist. 
Moreover, state-of-the-art techniques \cite{Wang2021gsnet} \cite{ZhouZhengyang2020RAMC}\cite{BaoHan2023Ssmf}\cite{doi:10.1137/1.9781611977172.38}\cite{yuan2018hetero} either rely on spatial features or use a spatial ensemble with pre-selected partitions. \textbf{Space partition} refers to dividing a geographical area into smaller regions for the purpose of modeling and prediction. Their partitioning process is completely separate from the model training process, thus the partitioned sub-regions obtained may not necessarily contribute to reducing the error in all areas. Moreover, pre-defined partitions may not be spatially homogeneous in terms of learning predictors of accident risk. This limits their accuracy and their ability to generalize to other regions without carefully redesigning the pre-selected regions. Thus, model performance is unsatisfactory and influenced by the user's external knowledge.

In this paper, We propose \textbf{LISA}, \underline{\textbf{L}}earning-\underline{\textbf{I}}ntegrated \underline{\textbf{S}}pace p\underline{\textbf{A}}rtitioning framework, to learn space partitions and spatial ensembles by addressing the heterogeneity problem. LISA is a generic framework and can work with various deep neural network implementation that has the ability to handle spatiotemporal features. Our code is available at https://github.com/BANG23333/LISA \underline{Our main contributions are summarized below}:

\begin{itemize}
\item We propose an integrated hierarchical framework to unify partitioning, training, and knowledge transfer. Our proposed method simultaneously obtains the partitions, while the models are learned.
\item We propose a novel fully automated partitioning algorithm, which guides the creation of the partitions, based on continuously reducing the prediction error.
\item We perform extensive experiments on a real-world accident dataset in the state of Iowa to demonstrate the effectiveness of the LISA framework working with various deep learning networks.

\end{itemize}

\section{Related Work}

Traffic accident analysis and forecasting have been studied during the past few decades. Early works used straightforward classic machine learning methods to address the accident prediction problem. Chong \cite{Chong2004TrafficAA} formulated a classification problem and used a decision tree and an artificial neural network on a dataset with police accident reports. In a similar setup, \cite{Lin2015} applied a random forest ensemble, K nearest neighbor, and Bayesian network to predict accidents. Abellán \cite{ABELLAN20136047} utilized probabilistic neural networks to address this problem. In a different setup, some works formulated a regression problem aiming to predict the frequencies of traffic accidents given some locations in a future period. ARIMA \cite{BarbaLida2014SSCw} and multinomial regression models \cite{CALIENDO2007657} were used to predict time-series changes of the accidents on road segments. Radwan \cite{ABDELATY2000633} adopted a negative binomial modeling technique to illustrate the inherent relationships between accidents and surrounding traffic conditions. Unfortunately, most classic machine learning methods have difficulty in capturing the complicated patterns of traffic accidents in heterogeneous space. 

With the availability of massive traffic, weather, and human activity data in recent years, it has been shown to be effective to solve traffic-related forecasting problems by using deep learning approaches \cite{JiangJiawei2024PPDD}\cite{HuMingzhi2023SERo}\cite{10.5555/3666122.3666555}\cite{10.1145/3474837}\cite{li2018diffusion}. Some researchers have attempted to introduce deep learning techniques on traffic accident forecasting problems. For example, LSTM \cite{Ren2018ADL}, attention-based \cite{Ren2018ADL}, and autoencoder-based \cite{Chen2018SDCAESD} are used to capture temporal patterns. Moreover, Najjar \cite{najjar2017combining} use CNN layers to learn spatial relationships.  Although, spatial-temporal deep learning methods bring some improvements, simply applying existing models to this problem results in unsatisfactory accuracy due to unique patterns of traffic accident occurrences. Zhou \cite{9242313} proposed a differential time-varying graph convolution network to dynamically capture traffic variations and boost accident prediction. Similarly, Wang \cite{Wang2021gsnet} proposed a GSNet with a geographical module and a weighted loss function to capture semantic spatial-temporal correlations among regions and solve data sparsity issues. However, the mentioned methods do not fully address the heterogeneity issue, as they rely only on spatial features or focus on a small-scale homogeneous region. 

To address the spatial heterogeneity issue, the deep learning framework with space partitioning ensembles has been proven to be a promising approach for other problems such as pandemic modeling \cite{9679027} and agriculture field classification \cite{10.1145/3474717.3483970}. Notably, Yuan \cite{yuan2018hetero} proposed Hetero-ConvLSTM to address the spatial heterogeneity challenge of the traffic accident forecasting problem by an ensemble of predictions from models learned from 21 pre-selected sub-regions. Similarly, An \cite{doi:10.1137/1.9781611977172.38} proposed HintNet partition the study area into levels of sub-regions based on historical accident frequencies. Nevertheless, those methods do not accurately capture underlying spatial heterogeneity as a result of pre-defined partitions. To tackle those challenges, we propose LISA, a novel learning-integrated space partitioning framework to automatically generate partitions while learning models and reducing prediction errors.

\vspace{-0.1in}

\section{Preliminaries}

\label{overall}

In this section, we describe the data source, definitions, feature extraction, and problem formulation for traffic accident forecasting.

\subsection{Data Source}
The data are collected in the state of Iowa from the year 2016 to the year 2018 from the following sources: \textbf{(1) Vehicle Crash Records} obtained from the Iowa Department of Transportation(DOT) \cite{iowadot} consist of the locations and time of vehicle crashes. \textbf{(2) RWIS (Roadway Weather Information System)} \cite{Mesonet} is an atmosphere monitoring system with 86 observation stations near state primary roads. \textbf{(3) COOP (National Weather Service Cooperative Observer Program)} is maintained by National Weather Service \cite{Mesonet} to monitor weather information such as precipitation, snowfall, and snow depth.\textbf{(4) Point of Interests (POI) data} are collected from HERE MAP API \cite{iowaheremap}. \textbf{(5) Iowa Road Networks data} \cite{Mesonet} contains basic road information such as speed limit and estimated annual traffic volume. \textbf{(6) Iowa Traffic Camera Data} is collected from 128 camera stations \cite{Mesonet} along state roads monitoring real-time traffic conditions.

\subsection{Definitions and Feature Extraction}
A $spatial$-$temporal$ $field$ $L \times T$ is a three-dimensional matrix, where $T = \{t_{1}, t_{2},...,t_{n}\}$ is a study period divided into equal length intervals (e.g., hours, days) and $L = \{l_{1}, l_{2},...,l_{m}\}$ is a two-dimension spatial grid partitioned from the study area.

\begin{definition}
\label{mmsk}
\textbf{Road Network Mask Map} $H$ \textit{is a binary mask layer generated by mapping the road network with primary and secondary roads onto the grid, where}
\begin{equation}
    \begin{aligned}
    H(l) = \begin{cases}1, & \textit {if there is at least one road segment in $l$}   \\
    0 & \textit {Otherwise}\end{cases}
    \end{aligned}
\end{equation}
\end{definition}

\begin{figure}[ht] \hspace*{0cm}
 \centering
 \includegraphics[width = 0.45\textwidth]{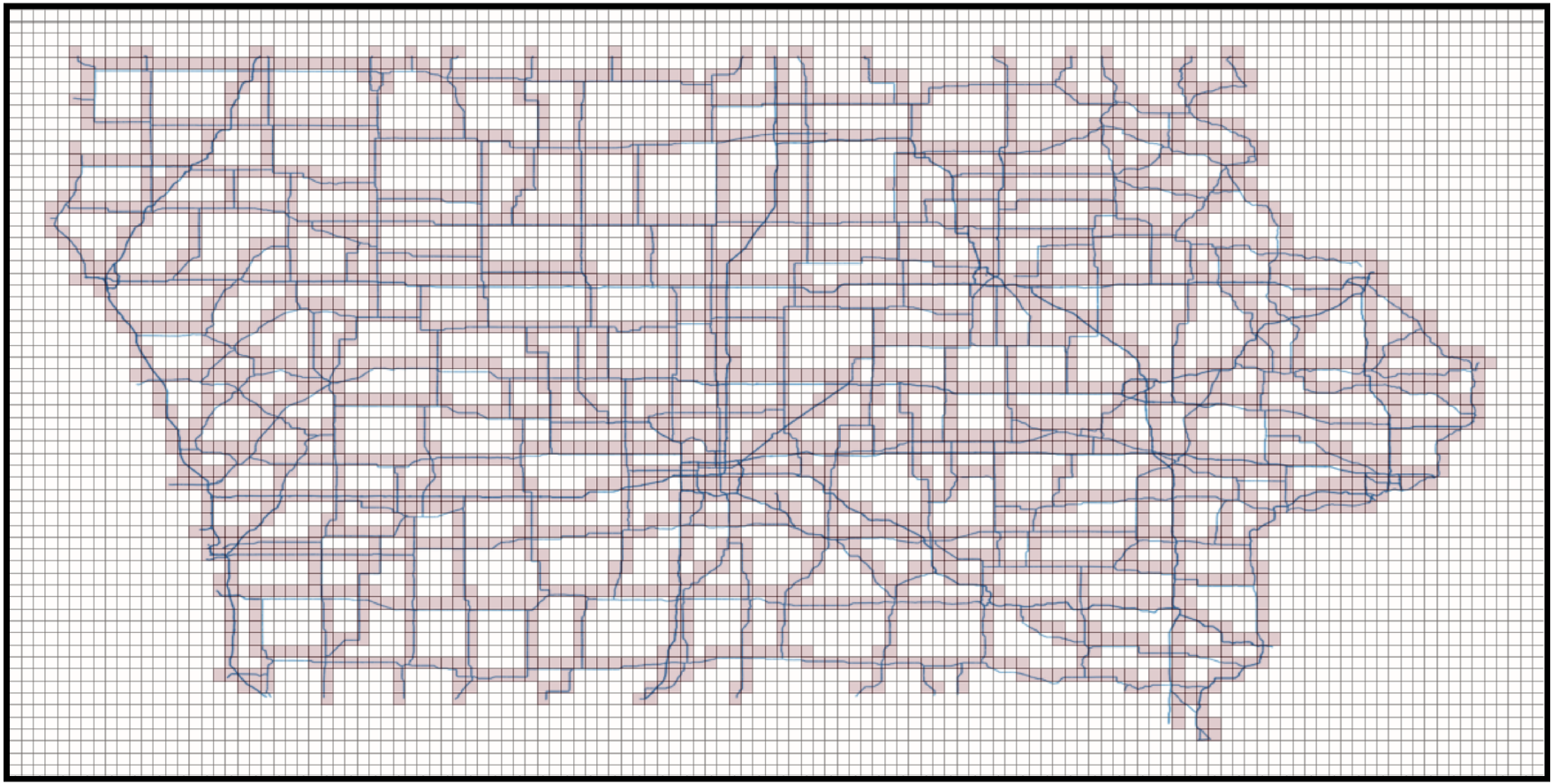}
 \caption{Grid on the state of Iowa and mask map with road networks. Red cell $l$ indicates $H(l)$ is 1.}
 \label{fig: grids}
\end{figure}

\begin{definition}
\textbf{Spatial Heterogeneity} \textit{is non-stationary of instances in a space \cite{10.1145/3161602}, which implies that the data distribution and spatial relationship with covariates varies over space. For example, the learned function to predict accidents might be very different from urban areas to rural areas.}
\label{SH}
\end{definition}

\begin{definition}
\textbf{Temporal Features} $F_T$ \textit{include the day of the week, day of the year, the month of the year, whether it is a weekend, and whether this is a holiday. Note $F_T$ remains unchanged across different locations s.}
\end{definition}
\begin{definition}
\textbf{Spatial Features} $F_S$ \textit{include the POI, road network information, and generated spectral features\cite{yuan2018hetero}. Note $F_S$ remains the same through time intervals t.}
\end{definition}
\begin{definition}
\textbf{Spatio-Temporal (ST) Features} $F_{ST}$ \textit{include weather information and real-time traffic conditions. $F_{ST}$ varies over locations and time.}
\end{definition}

\begin{definition}
$\textbf{Traffic Accident Occurrence Tensor \emph{O(l, t)}}$ \textit{is used to denote the number of accidents that occurred in grid cell $l$ during time $t$.}
\end{definition}

\noindent\textbf{Feature Summary:} Our framework deals with three types of input features: spatial, temporal, and spatiotemporal. In this work, we extract 47 features as the input, including 5 temporal features(e.g., day of year, holiday), 29 spatial features (e.g., total road length, avg. speed limit), and 13 spatiotemporal features (e.g., traffic volume) for each location $l$ and time interval $t$.

%\subsection{Problem Statement}
%\textbf{Given} the socio-environmental features $F$ and accident frequencies $O$ for all the locations in time window $\{t_1,t_2,.., t_n\}$, the traffic accident forecasting problem is to \textbf{Predict} accident frequencies in future one time interval $t_{n+1}$ for all locations $l_m \in L$. The \textbf{objective} is to minimize prediction errors. As a basic assumption of our problem, all traffic accidents occur along the road system and spatial heterogeneity exists in the data. This paper uses seven consecutive days of data to predict the eighth day ($n=7$).

\subsection{Problem Formulation}
The problem is formally defined as:\\ 
\textbullet\ \textbf{Given:}\\
\indent \textendash\ A spatial-temporal field $L \times T$\\
\indent \textendash\ A road network\\
\indent \textendash\ A list of accidents with time and locations\\
\indent \textendash\ A set of feature data \\
%\textendash\ A road network mask map $M$ \\
%\textendash\ Traffic Accident Occurrence Tensor $O$ for a time window $[t-n, t-1]$ for all $l\in L$, $n<t$\\
%\textendash\ A set of feature tensors $F = \{F_T, F_S, F_{ST}\}$ for the same time window for all the locations $l\in L$\\
\textbullet\ \textbf{Find:}\\
\indent \textendash\ Predicted accident count in every $l\in l$ for $t$: $\hat{O}(l,t)$\\
\textbullet\ \textbf{Objective:}\\
\indent \textendash\ Minimize the prediction error\\
\textbullet\ \textbf{Constraints:}\\
\indent \textendash\ All traffic accidents occur along road system.\\
\indent \textendash\ Spatial heterogeneity exists in the data.\newline

\noindent where $n$ is the time length of the input for each prediction and $t$ is a single day. In this paper, we use seven consecutive days of data to predict for the eighth day ($n=7$).

\section{Methodology}
\label{solu}
The proposed LISA framework consists of two main nested modules: i) Integrated Hierarchical Partitioning Training (I-HPT), and ii) Partition Learner (PL). The PL module is nested inside the I-HPT, where the I-HPT utilizes PL to generate partitions and corresponding trained models given historical extracted features. The LISA framework can work with various deep-learning implementations that handle spatiotemporal inputs.

\subsection{Integrated Hierarchical Partitioning Training}
%Contrary to our prior work HintNet, in which the partition process was completely separate from the training process, the proposed method partitions the area and learns the models concurrently. 
To reduce spatial heterogeneity in each partition while considering model performance, we develop a neighbor-sampling mechanism to gradually form a partition, where prediction accuracy is used to guide the partitioning process. In the rest of this section, we first introduce the architecture of I-HPT and then explain the PL module.

\begin{figure*}[t]
 \centering
 \includegraphics[width = 1.0\textwidth]{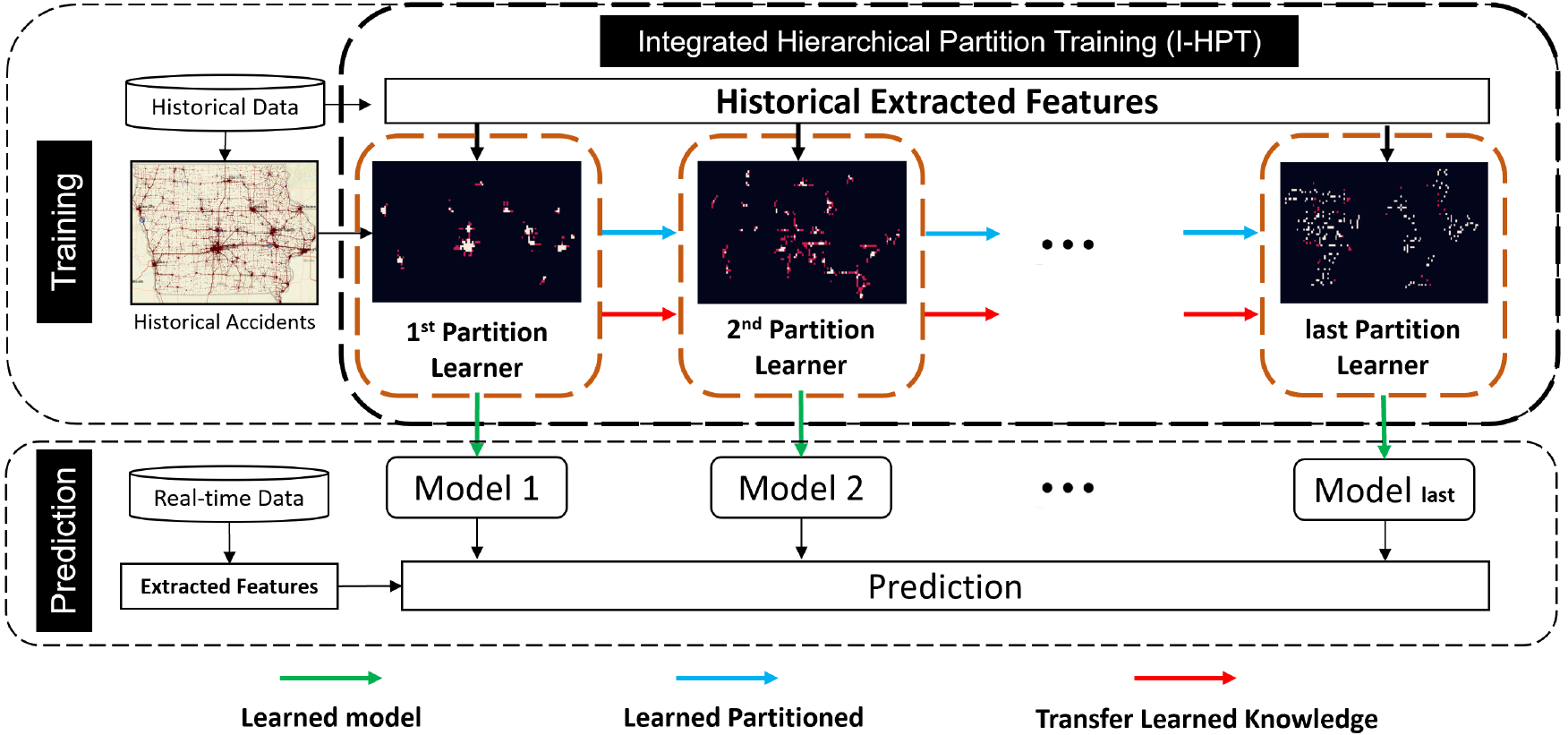}
 \caption{The overall architecture of Partition Leaner (PL) module.}
 \label{fig: PL}
\end{figure*}

\begin{figure*}[t]
 \centering
 \includegraphics[width = 0.9\textwidth]{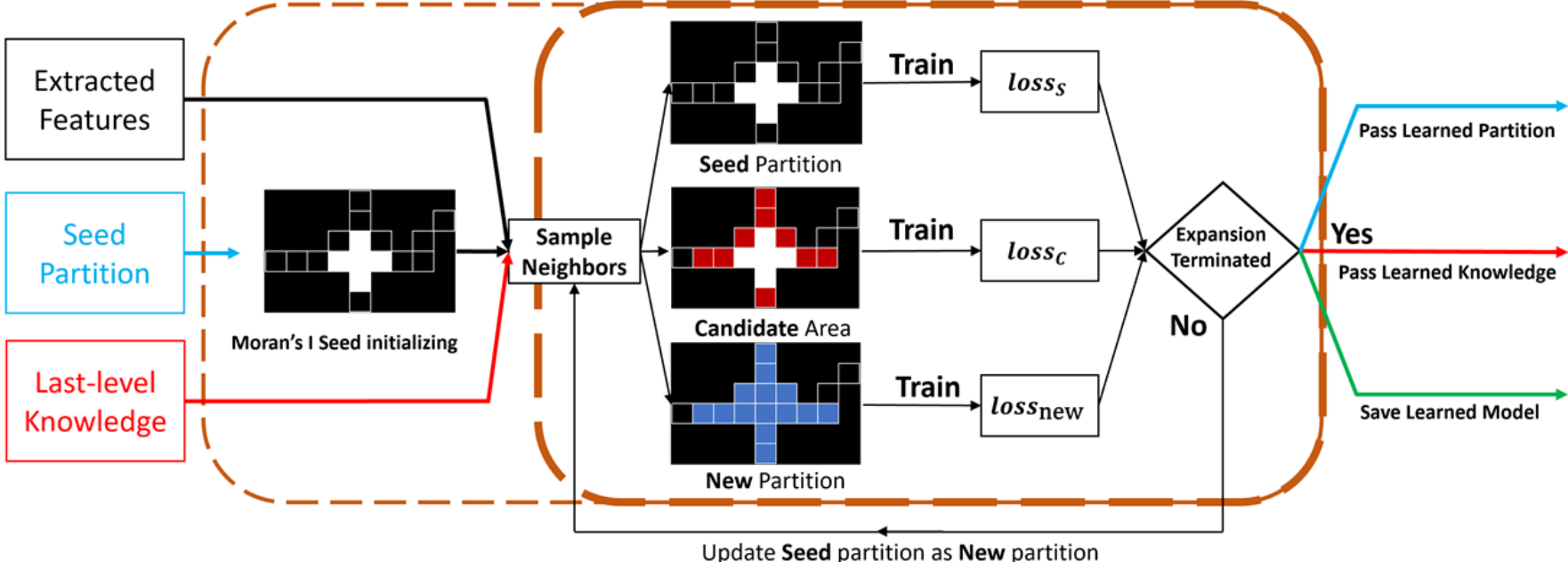}
 \caption{The overall architecture of I-HPT (Brown box represents Partition Learner module).}
 \label{fig: framework}
\end{figure*}

The procedure of I-HPT is shown in Figure \ref{fig: framework} and 
%Meanwhile, the model parameters learned from earlier partitions are transferred to initialize model parameters in the next partition to further smooth the training process.
Algorithm \ref{alg. Multi-SP}. Given a deep learning model (e.g., CNN, ConvLSTM) with initial parameters $\theta$, the extracted features $F$ are fed into I-HPT, and the Partition Learner (PL) is used to iteratively generate partitions from the study area based on the current model's training error distribution and use the training data from the new partition to update the model. 

\begin{algorithm2e}
	\SetKwInput{Input}{Input}
	\SetKwInput{Output}{Output}
	\LinesNumbered
	\DontPrintSemicolon
	\BlankLine
	\caption{Integrated Hierarchical Spatial Partition Training (I-HPT)}
	\label{alg. Multi-SP}
	%\begin{algorithmic}[1]
	\Input{ study area $L$, accident tensor $O$, extracted $F$, radius $r$, tolerance $\gamma$} 
	\Output{partitions $P$, trained models $\theta$}
	\BlankLine
	
    initialize model parameters $\theta_{current}$ \;
    initialize $P$ and $\theta$ as empty lists \;
	\While{$L$ is not empty}{

    $P_{current}$, $\theta_{current}$ = \textbf{PL}($L$, $O$, $F$, $\theta_{current}$, $r$, $\gamma$)	\;
    $P$.append($P_{current}$)\;
    $\theta$.append($\theta_{current}$)\;
    $L$ = $L$ - $P_{current}$ \;
	}
	
	\textbf{return} $P$, $\theta$
\end{algorithm2e}

The inputs are the study area, historical accident occurrence tensor, extracted features, and two parameters including radius $r$ and tolerance $\gamma$. parameters $r$ and $\gamma$ are used in the PL module. In Line 1 and Line 2, algorithm 1 starts with a randomly initialized deep learning model $\theta_{current}$. Then, two empty lists $P$ and $\theta$ are created to store obtained partitions and trained models. From Line 3 to Line 7, partitions and models are generated hierarchically using PL (Algorithm \ref{alg. pl}) given the remaining study area, accident tensor, extracted features, the current model, and two parameters. Meanwhile, the current partition and model are added to the list $P$ and list $\theta$ separately in each iteration. In Line 7, the partitioned region is removed from the remaining study area $L$, thus $L$ keeps shrinking in each iteration as partitions are formed. This process is terminated by a designed error-based tolerance mechanism discussed later or until all study area is explored. Intuitively, each partition tends to represent similar accident patterns, compared to the rest of the region, which has heterogeneity.

\subsection{Partition Learner (PL) module}
\label{sec: PL}
The challenge arises when trying to identify the appropriate partition in each iteration. As previously mentioned, it is beneficial to partition a less-heterogeneous sub-region, but less heterogeneity cannot promise models that perform best in terms of getting lower validation errors on the whole study area. Therefore, the PL module is designed to form a comparatively homogeneous sub-region and consider training performance simultaneously. In the rest of this section, we will discuss Local Moran's I test, a statistical test employed in the partitioning process and then explain the partitioning and training process. 

\textbf{Local Moran's I test} and \textbf{Local Geary's C} belong to the family of local spatial association indicators \cite{https://doi.org/10.1111/j.1538-4632.1995.tb00338.x}, and they are commonly used in understanding spatial heterogeneity\cite {LaohasiriwongWongsa2018Acos}. In our case, two tests on each location indicate the extent of significant spatial clustering of similar accident occurrences around that location, which indicates the degree of spatial heterogeneity. They are good at detecting significant spatial clustering at a local level and identifying outliers\cite{https://doi.org/10.1111/j.1538-4632.1995.tb00338.x}. Note that the spatial heterogeneity function can be replaced by other methods based on different problems.

The local Moran's I value is defined as
\begin{equation}
I_i =  Z_i \sum_{j}W_{ij} Z_j,
\end{equation}
where $I_i$ is the local Moran Statistic for location $i$, and $Z_i$ = $y_i$ - $\Bar{y}$ is the deviation of accident count between $y_i$ in location $i$ and the mean occurrence $\hat{y}$ of the whole study area. Location $j$ is the neighbor of location $i$ if they are adjacent in a queen's move. $W_{ij}$ is the adjacency weight matrix, and $W_{ii}$ = $0$. Note that the term, location, refers to the grid cell in our problem setting. The significance of $I_i$ is also tested under $H_0:$ \textit{no spatial heterogeneity exists} against $H_1:$ \textit{spatial heterogeneity exists.}

Similarly, the local Geary's C value is defined as
\begin{equation}
C_i =  \frac{1}{m_2}\sum_{j}W_{ij} (y_i - y_j)^2,
\end{equation}
where $m_2 = \sum_{i}(y_i - \Bar{y})/n$, n is the number of neighbors. Both methods compare the difference between neighboring values. Local Geary's C uses squared distance, and local Moran's I uses standardized covariance. We tested both measurements in the experiments.

\textbf{The procedure of PL} is illustrated in Figure \ref{fig: PL}. Starting from the left part of the figure, the first step is to utilize the Local Moran's I test to initialize a group of grid cells with significant positive local Moran's I scores under the hypothesis test that represents the regions with similar neighboring accident occurrences, where their accident frequencies are significantly higher than rest of study area. However, a selected significantly homogeneous region might not be the best partition in terms of training models and reducing errors. Therefore, in the next step, PL learns and expands based on the seed region step by step while monitoring local training errors and terminating if errors surge. The second step is shown on the right side of the diagram. The seed partition is expanded step by step by sampling neighbors while three models are trained on the seed partition, candidate area, and new partition respectively. Note that the new partition is the concatenation of the seed partition and candidate area. In Figure \ref{fig: PL}, the seed partition, candidate area, and new partition are shown as white region, red region, and blue region on the map. Finally, the PL module repeats this process until the prediction error on the new partition exceeds the sum of errors on both the seed partition and candidate partition is greater than a tolerance $\gamma$. The intuition of designed error-based termination is that we assume multiple models perform better than a single model on a heterogeneous area based on definition 2 and observations from related studies \cite{10.1145/3161602}\cite{9679027}\cite{10.1145/3474717.3483970}\cite{yuan2018hetero}. Therefore, this expanding process will terminate if we find two models (seed model and candidate model) that combined perform better than a single model (new model) in the same area because it indicates that the new area becomes heterogeneous. However, as the deep neural network is non-deterministic, we relax this assumption and design a tolerance mechanism to attempt sampling more neighbors to check if better results are possible. The tolerance mechanism keeps the prior best model if attempts fail.

\textbf{Algorithm \ref{alg. pl}} shows the details of PL. The inputs are the remaining study area, historical accident tensor, extracted feature, two parameters, and the previous model from the last iteration. The outputs are a partitioned sub-region and a corresponding trained model. We begin with introducing the \textbf{SampleCandidates()} function and then explain Algorithm 2 step by step. Lastly, we propose a novel heuristic search function called \textbf{SpatialGradientSearch()} to replace SampleCandidates() and further improve the partitioning.

\textbf{SampleCandidates()} is a basic search function and used to sample candidate neighboring locations given a predefined searching range $r$. Specifically, for grid cells in the seed region, this function selects all cells within a Manhattan distance smaller than radius $r$ from the remaining study area excluding the seed region. The selected grid cells output as the candidate area and the region combined with both candidate area and seed area is the output of a new partition. We only sample grid cells with positive Local Moran's I score because those cells are relatively homogeneous to the seed region. In addition, sampled candidates might be insufficient to train a model if the size of the seed region is too small. Thus, we designed a re-sampling strategy that keeps sampling nearby neighbors until the total samples included are greater than a threshold multiplied by the model batch size. This threshold is a multiplier of the batch size and controls the minimum candidate samples returned in this function.

\begin{algorithm2e}
	\SetKwInput{Input}{Input}
	\SetKwInput{Output}{Output}
	\LinesNumbered
	\DontPrintSemicolon
	\BlankLine
	\caption{Partition Learner (PL)}
	\label{alg. pl}
	%\begin{algorithmic}[1]
	\Input{remaining study area $L$, accident tensor $O$, extracted feature $F$, previous model $\theta_{prev}$, radius $r$, tolerance $\gamma$} 
	\Output{partition $P_{best}$, trained model $\theta_{best}$}
	\BlankLine
	
    Initialize Seed Partition $P_s$ = LISA\_test($L$, $O$) \;
    $\theta_{best}$, $P_{best}$ = \textbf{Train}($P_s$, $F$, $\theta_{prev}$) \;
    Set $ctr$ as $0$ \;	
    \While{$P_s \not = L$}{
        $C$, $P_{new}$ = \textbf{SampleCandidates}($P_s$, $\theta_{best}$, $r$)  \;
        $\theta_s$, $loss_s$ = \textbf{Train}($P_s$, $F$, $\theta_{prev}$) \;
        $\theta_c$, $loss_c$ = \textbf{Train}($C$, $F$, $\theta_{prev}$) \;
        $\theta_{new}$, $loss_{new}$ = \textbf{Train}($P_{new}$, $F$, $\theta_{prev}$) \;
        $P_s = P_{new}$  \;
        \If{$loss_{new} > loss_s + loss_c$}
        {
        $ctr ++$
        }\Else{
        reset $ctr$ as $0$ \;
        $\theta_{best} = \theta_{new}$  \;
        $P_{best} = P_{new}$  \;
        }

        \If{$ctr >= \gamma$}
        {
        $\mathbf{break}$
        }
    }	

	\textbf{return} $P_{best}$, $\theta_{best}$
\end{algorithm2e}

Algorithm~\ref{alg. pl} takes the following steps:

1) PL module starts by initializing a seed partition $P_s$ using the Local Moran's I test from the remaining study area $L$ in Line 1. Then, the final output variables $\theta_{best}$ and $P_{best}$ are initialized by training on the initial seed partition. %thus the smallest possible partition PL produces is the initial seed partition. It will be updated later if the algorithm decides to expand. 
Lastly, tolerance counter $ctr$ is set to 0.

2) The iterative partitioning and training part starts from Line 4. Specifically, given the current seed partition, \textbf{SampleCandidates} function is used to select candidate area in Line 5. Next, the seed model, candidate model, and new model are learned from their corresponding regions from Line 6 to Line 8. Meanwhile, their total validation errors are obtained as $loss_s$, $loss_c$, and $loss_{new}$. Lastly, the initial partition is updated by the new partition in Line 9. 

3) The error-based termination is from Line 10 to Line 17. In each iteration, if the total validation error from the new model is greater than the sum of total validation errors from both the initial and candidate models, the counter will be incremented by one. Otherwise, the counter is reset as zero, and $\theta_{best}$ and $P_{best}$ are updated (Line 13 to Line 15). These variables are updated only if better results are found. However, if the algorithm cannot find a better partition within $\gamma$ number of samplings, the loop will be terminated at Line 17. Finally, the saved best partition and its corresponding trained model are returned.

\textbf{Framework Inference on partitions} In Algorithm \ref{alg. Multi-SP}, we learn models on each partition. In the prediction phase, we use learned models to make predictions on their corresponding partitions. All predictions are mapped to the original study area at the end.

\subsection{Spatial Gradient Search}

\begin{figure}[ht] \hspace*{0cm}
 \centering
 \includegraphics[width = 0.47\textwidth]{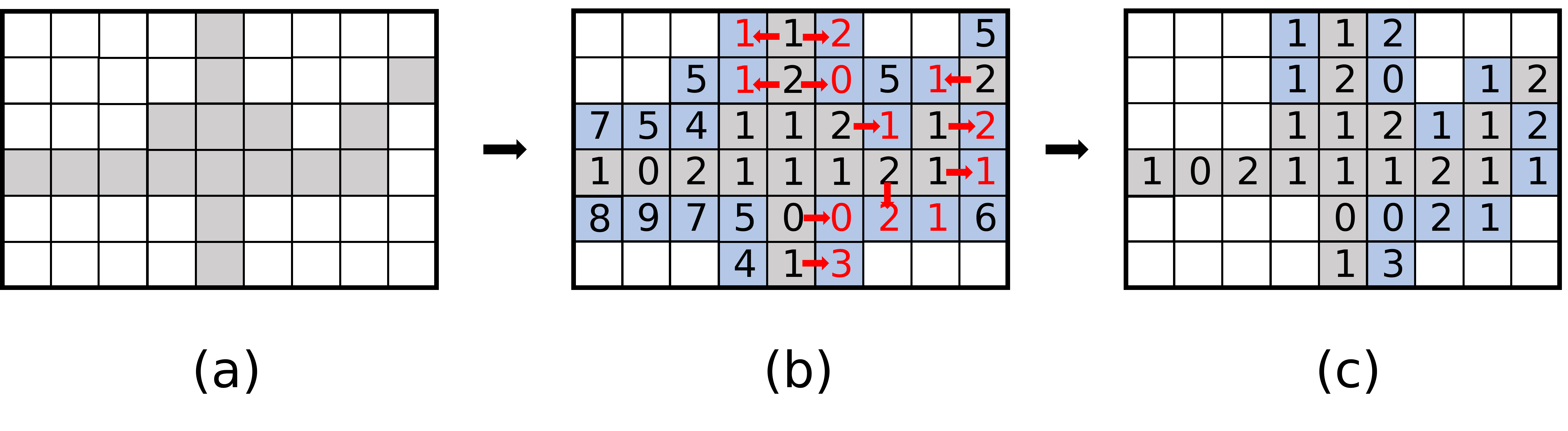}
 \caption{An example of Spatial Gradient Search (The grey area is the seed region, and the blue area is the candidate region to be expanded. The numbers are the losses made by the model learned from the seed region.)}
 \label{fig: SGS}
\end{figure}

\textbf{SpatialGradientSearch()} is an alternative search function and proposed to select homogeneous neighboring locations given a seed region and corresponding trained model $\theta$ on the seed region. \textbf{SpatialGradientSearch()} can directly replace \textbf{SampleCandidates()} in the line 5 of Algorithm. 2. To identify a homogeneous region from surrounding locations, the key idea is to use the trained model on the current region to validate its performance in surrounding locations. Those locations with fewer errors are relatively homogeneous to the existing partition. To achieve this goal, we design a Spatial Gradient Search function to select locations based on gradient-based sorting over training samples of each location. Take a toy example in Figure. \ref{fig: SGS}, SpatialGradientSearch() searches for surrounding relative homogeneous area from image (a) to image (c). Given the definition of heterogeneity in Definition \ref{SH}, we assume that the model trained on the grey seed region should perform similarly on a homogeneous region. Therefore, we validate the model on surrounding locations and select top $K$ locations with fewer errors made by the existing model, so candidates with homogeneous patterns can be selected. Specifically, in the first line of Algorithm \ref{alg. SGS}, SampleCandidates() returns candidate locations $C$ within a Manhattan distance smaller than radius $r$ from the remaining study area excluding the seed region. To select homogeneous locations from the candidates, from line two to line three, we use a trained model from the seed region to infer $\hat{Y}_C$ on the candidate area, and losses are computed on each location. To capture spatial correlations, an average convolution $\psi$ is used to smooth the loss given kernel size $r$. Lastly, we identify the top $K$ locations with the least errors to be used in the next expansion. The selected grid cells output as the candidate area and the region combined with both candidate area and seed area is the output of a new partition.

\begin{algorithm2e}
	\SetKwInput{Input}{Input}
	\SetKwInput{Output}{Output}
	\LinesNumbered
	\DontPrintSemicolon
	\BlankLine
	\caption{Spatial Gradient Search}
	\label{alg. SGS}
	%\begin{algorithmic}[1]
	\Input{ seed area $P_s$, seed model $\theta$, radius $r$} 
	\Output{selected candidate area $C'$, partition $P_{new}$}
	\BlankLine
	
    %initialize model parameters $\theta_{current}$ \;
    %initialize $P$ and $\theta$ as empty lists \;

    $C$ = \textbf{SampleCandidates}($P_s$, $L$, $r$) \;
    $\hat{Y}_C = f_{\theta}(F_{C}) $ \;
    $Loss_C$ = $\frac{1}{T} \sum^T_{t=1} (Y_C^t - \hat{Y}_C^t)^2$ \;
    $Loss_C'$ = $convolution\Psi(Loss_C)$ \;
    $indices = Argmin(Loss_C', K)$ \;
    $C'$ = $C$[$indices$] \;
    $P_{new}$ = $C'$ + $P$
	%\While{$L$ is not empty}{

    %$P_{current}$, $\theta_{current}$ = \textbf{PL}($L$, $O$, $F$, $\theta_{current}$, $r$, %$\gamma$)	\;
    %$P$.append($P_{current}$)\;
    %$\theta$.append($\theta_{current}$)\;
    %$L$ = $L$ - $P_{current}$ \;
	%}
	
	\textbf{return} $C'$, $P_{new}$
\end{algorithm2e}

\subsection{Optimization} 
In this problem, We use the mean square error (MSE) loss to measure the performance of our model as we predict the frequency of future accidents. MSE is widely accepted as the loss function in the traffic accident forecasting problems. LISA framework can also work with other loss functions such as cross-entropy depending on the targeting problem. The loss function is denoted as:
\begin{equation}
Loss =  \frac{1}{T} \sum^T_{t=1} (Y_t - \hat{Y}_t)^2,
\end{equation}
where $Y_t$ is the ground truth and $\hat{Y}_t$ is the predicted values of all grid cells at time interval $t$.\\ 

\textbf{Complexity Analysis} Assuming the cost of computing the baseline network used in the LISA framework is $O(Ln)$, where $L$ is the number of grid cells in the study area and $n$ is the training cost for each grid cell. In the best case, the computing cost of LISA is $O(Ln)$, where the initial seed area covers the whole study area and no expansion is needed. In the worst case, LISA needs to expand from the smallest seed region to cover all grid cells, and LISA needs to re-train models at every step. The number of covered cells increases from $1$ to $L$, thus the computing cost is $O(\frac{L(L+1)}{2}n)$, therefore we have the complexity of $O(L^2n)$ for a large model size.

%\begin{algorithm2e}
	%\SetKwInput{Input}{Input}
	%\SetKwInput{Output}{Output}
	%\LinesNumbered
	%\DontPrintSemicolon
	%\BlankLine
	%\caption{Heuristic Religion Expanding}
	%\label{alg. expand_alg}
	%%\begin{algorithmic}[1]
	%\Input{ study area $L$, seed area $S$, seed model $\theta_{seed}$, extracted feature $F$, threshold %$\gamma$} 
	%\Output{candidate area $C$, trained models $\theta_{new}$}
	%\BlankLine
%	
%    $P_{new}$ = \textbf{SampleCandidates}($P_s$, $L$, $r$)  \;
%    $P = []$ \;
%	\For{$l$ in $P_{new}$}{
%%        $\hat{Y_l} = f_{\theta_{seed}}(F_{l})$ \;
%        $loss_{l} = \frac{1}{T} \sum^T_{t=1} (Y_l - \hat{Y_l})^2$ \;
%        Compute gradient $\nabla f(\theta_{seed})$ by $loss_{l}$ \;
%        \If{$\nabla f(\theta_{seed})$ < $\gamma$}{
%            $P.append(l)$
%            }
%	}
%	
%	\textbf{return} $P$, $\theta$
%\end{algorithm2e}

\section{Experiment}
\label{eva}

\subsection{Experiment Settings}
\subsubsection{Data Preprocessing}
The data are collected in the state of Iowa from the year 2016 to the year 2018. The data from the years 2016 and 2017 are used as the training set, and $20\%$ of the training set is randomly selected as a validation set. The data from the year 2018 is used as a testing set. The state of Iowa area is partitioned by $5$ km $\times$ $5$ km square cells and converted to a grid with the size of $128 \times 64$.

\subsubsection{Evaluation Goals}
We wish to answer the following questions in the experiments: (1) Does the proposed framework outperform baselines in areas with different levels of heterogeneity? (2) How do parameter settings influence model performance? (3) How do different feature groups contribute to the performance of methods in the areas with different levels of heterogeneity? %(4) How does the proposed method perform on different partitions? 
(5) Do the predictions make sense and stay consistent with the ground truth?

\subsubsection{Metrics}
We evaluate the model performance by measuring the mean squared error (MSE). Moreover, we use the Cross-K function \cite{doi:https://doi.org/10.1002/9781118445112.stat07751} with Monte Carlo simulation to evaluate the spatial correlation between predictions and ground truth in the masked study area.

\subsubsection{Parameter Configurations}
The proposed method is trained by minimizing the mean squared error by using the Adam optimizer \cite{KingmaDiederikP2014AAMf} with settings $\alpha=10^{-4}$, $\beta_1=0.9$, $\beta_2=0.999$, maximum training iteration = $20$, and $\epsilon=10^{-8}$. An early stopping mechanism is employed while training models, and the training process is terminated if the validating loss stops decreasing for 5 consecutive epochs. Normalization is used to transform data into the range $[0, 1]$. 
% Regarding parameter setting of radius and tolerance, we test the multiplier in SampleCandidates() function with values $20$, $25$, $30$, $35$, $40$, radius with Manhattan distance $2$, $3$, $4$, $5$, $6$, and tolerance with $0$, $1$, $2$, $3$, $4$. The results are presented in the ablation study section.

\subsubsection{Platform}
We run the experiments on High-Performance Computer System with Intel Xeon E5 2.4 GHz and 256 GB of Memory. We use a GPU node with Nvidia Tesla V100 Accelerator Cards with the support of Pytorch library \cite{NEURIPS2019_9015} to train the deep learning models.

\subsubsection{Baselines}
We compare the proposed method with the following baselines.

\textbf{(1) Historical Average (HA)} is the historical average daily accident counts calculated from the training set. 

\textbf{(2) Least Squares Linear Regression (LR)} \cite{MontgomeryDouglasC.2012Itlr} is a classic machine learning algorithm for estimating linear relationships between predictors and response variables.

\textbf{(3) Decision Tree Regression (DTR)} by \cite{RyzinJohnVan1986CaRT} is the regression tree, and the maximum depth is set to 30.

\textbf{(4) Long Short Term Memory (LSTM)} \cite{HochSchm97} a recurrent neural network architecture with feedback connections. We stack two fully-connected LSTM layers.

\textbf{(5) ConvLSTM} \cite{ShiXingjian2015CLNA} is a recurrent neural network with convolution layers for spatial-temporal prediction. We use a two-layer network and set the number of hidden dimensions as several features.

\textbf{(6) DCRNN} \cite{li2018diffusion} is an advanced traffic forecasting model which captures the spatial dependency using bidirectional random walks on the graph and the temporal dependency using the encoder-decoder architecture.

\textbf{(7) GSNet} \cite{Wang2021gsnet} is a recent deep-learning method to learn geographical and semantic aspects for traffic accident prediction. We modify the weighted loss function to fit our problem. The loss calculated on each sample is weighted by a multiplier, 0.05, 0.2, 0.25, and 0.5 when its accident count is 0, 1, 2 to 4, and greater than 4.

\textbf{(8) Hetero-ConvLSTM (H-ConvLSTM)} \cite{yuan2018hetero} is an advanced deep learning framework to address spatial heterogeneity in the traffic accident forecasting problem. It applies multiple ConvLSTM on pre-defined sub-regions with size $32\times32$. We use the same parameter setting of ConvLSTM in this method as the ordinary ConvLSTM baseline.

\textbf{(9) HintNet} \cite{doi:10.1137/1.9781611977172.38} is a recent deep learning framework to partition study areas based on historical accident frequencies, and we use the same parameter settings.
When LISA is applied, only the network of HintNet is used in the experiments.

The LISA framework can work with various deep learning networks, and we choose four advanced deep learning networks including LSTM, ConvLSTM, DCRNN, and HintNet as build-in networks of LISA. We compare the performance between the original baselines and baselines improved by LISA. Moreover, two local spatial association indicators, Local Moran's I and Geary's C, are tested. We don't use Hetero-ConvLSTM as a built-in network because Hetero-ConvLSTM is built based on ConvLSTM.

\subsection{Prediction Accuracy Analysis}
In this section, we first compare the model performances in different regions and then analyze the spatial correlation between predictions and ground truth.

\subsubsection{Prediction Error Comparison Between Regions}

\begin{figure}[ht] \hspace*{0cm}
 \centering
 \includegraphics[width = 0.45\textwidth]{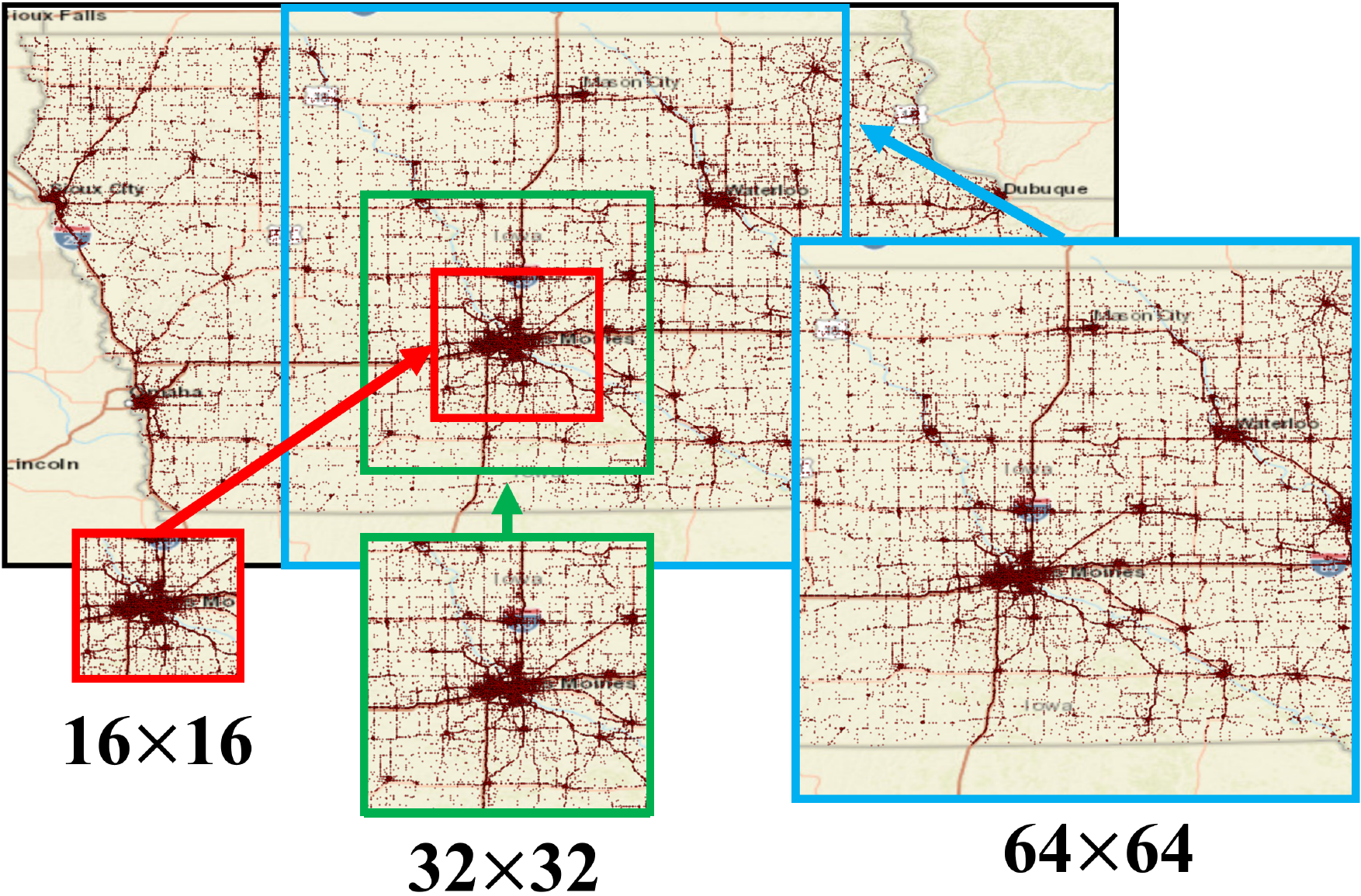}
 \caption{Four tested regions are indicated by colored boxes with grid size. The background is the plot of historical accident locations}
 \label{fig: comparision}
\end{figure}

To measure the performance of the proposed method and baselines on a different degree of spatial heterogeneity, we select 4 types of regions from the study area as demonstrated in Figure \ref{fig: comparision}. The red, green, blue, and black boxes on the study area represent homogeneous ($16\times16$), less-homogeneous ($32\times32$), heterogeneous ($64\times64$), very heterogeneous ($128\times64$) regions. Note that the red box covers the area of Des Moines, which is the capital city of the state of Iowa representing the most populated region. %In terms of implementation in four regions, the size of the convolution filter is set as $5\times5$ in the homogeneous and less-homogeneous region and $7\times7$ in the heterogeneous and very heterogeneous region for both our proposed method and ConvLSTM baselines. Moreover, it is computationally infeasible to apply GSNet on the largest region ($128\times64$) due to the oversized graph operations, thus we apply two independent GSNet models on the left and right halves of the study area and measure the total prediction error.

Table 1 demonstrates the comparison between the proposed method and baselines in regions with different degrees of spatial heterogeneity, where LISA can improve all baselines. LISA without $*$ uses local Moran's I test, and LISA with $*$ uses local Geary's C test instead. Linear regression performs stably in four tested regions, but its performance is only close to the result of the historical average. Interestingly, state-of-art methods such as ConvLSTM and GSNet have been influenced significantly by the changing degree of spatial heterogeneity. Both of them achieve decent results in the smaller homogeneous region but perform poorly in large and heterogeneous regions. Given baseline networks including LSTM, ConvLSTM, DCRNN, and HintNet, LISA improve the performance in all level of heterogeneity. LISA with local Moran's I test performs slightly better than LISA with local Geary's C test on different base-models. Results with underlines in Table 1 demonstrate that LISA can work with different existing networks and constantly improve the model performance by an average of $13.0\%$. Noticeably, LISA framework improves ConvLSTM and DCRNN by $43.1\%$ and $16.4\%$ respectively in most heterogeneous areas. LISA with HintNet network using local Moran's I test achieves the best results.

\begin{table}[t]
\label{tab: comp}
\begin{threeparttable}[b]
\vspace{-0.0in}
\caption{Performance Comparison}
\vspace{0mm}
\label{tab:performance}
\vskip 0.0in
\begin{center}
\begin{small}
\begin{sc}
\begin{tabular}{p{2.0cm}p{1.2cm}p{1.2cm}p{1.2cm}p{1.2cm}}
\toprule
\multirow{1}{*}{\thead{\textbf{\tiny{$\textbf{MSE}\times 10^{-3}$}}}} &

\footnotesize{\tiny{$128\times64$}} & \footnotesize{\tiny{$64\times64$}} & \tiny{\tiny{$32\times32$}} & \footnotesize{\tiny{$16\times16$}} \\
\midrule
HA   & 2.711 & 3.215 &  7.413 &  22.02  \\
LR   & 2.645 & 3.211 &  7.546 &  22.09  \\
DTR   & 3.329 & 8.242 &  13.329 &  38.86  \\

\footnotesize{H-ConvLSTM} & 2.449 & 2.826 &  8.108 &  19.72 \\
GSNet   & 3.673 & 3.635 &  8.002 &  17.88 \\

\midrule
LSTM   & 2.730 & 3.311 &  8.543 &  26.44  \\
LISA\tiny{-LSTM}$^*$   & 2.524 & 3.006 &  \underline{8.391} &  \underline{24.99}  \\
LISA\tiny{-LSTM}   & \underline{2.474} & \underline{2.913} &  8.414 &  25.22  \\

\midrule
ConvLSTM   & 3.917 & 4.420 &  8.103 &  19.74  \\
LISA\tiny{-ConvLSTM}$^*$   & 2.323 & 3.110 &  7.134 &  \underline{19.12}  \\
LISA\tiny{-ConvLSTM}   & \underline{2.229} & \underline{2.707} &  \underline{6.965} &  19.16  \\

\midrule
DCRNN   & 2.542 & 2.917 &  7.823 &  17.78 \\
LISA\tiny{-DCRNN}$^*$  & 2.196 & \underline{2.478} &  6.305 &  17.76 \\
LISA\tiny{-DCRNN}  & \underline{2.126} & 2.486 &  \underline{6.079} &  \underline{17.43} \\

\midrule
HintNet   & 2.111 & 2.535 &  6.029 &  17.44 \\
LISA\tiny{-HintNet}$^*$   & 2.020 & 2.393 &  5.828 &  17.42 \\
LISA\tiny{-HintNet}   & \textbf{1.971} & \textbf{2.288} &  \textbf{5.423} &  \textbf{17.41} \\

\hline

\bottomrule
\end{tabular}
\begin{tablenotes}
    \item $*$: Local Geary's C
\end{tablenotes}
%\begin{tablenotes}
%    \item [*] \textbf{LISA}   
%\end{tablenotes}
\end{sc}
\end{small}
\end{center}
\end{threeparttable}
\label{table: aaa}
\vspace{-0.2in}
\end{table}

\subsubsection{Cross K function with Monte Carlo Simulation}

In this part, we use the Cross-K function\cite{doi:https://doi.org/10.1002/9781118445112.stat07751} with Monte Carlo Simulation\cite{TaoRan2019FCKa} to evaluate the accuracy of predicted locations. The Cross-K function measures the spatial correlation between the predicted locations with accident happened and true locations. Specifically, we calculate the average density of predicted accidents within every distance $d$ of a true accident in each day as shown in equation 4:

\begin{equation}
\hat{K}(d) =  \lambda_j^{-1}\sum_{i \neq j}I(d_{ij} \leq d)/n,
\end{equation}

\noindent where $\lambda$ is global density of event $j$, and $I()$ is an identity function which equals one if real distance $d_{ij}$ is smaller than $d$, else equals zero. $n$ is the number of events $i$. In addition, the variance of the Cross-K function with complete spatial randomness can be estimated by Monte Carlo Simulation, thus we can assess the results of the cross-k function with the locations in the study area allowed by our mask map defined in Def. \ref{mmsk}. We found that LISA with the HintNet network performs best, and the curve of cross-K functions is higher than the curves generated from other baselines, which indicates that the LISA framework can significantly improve spatial correlation with ground truth. 

\begin{figure}[ht] \hspace*{-1cm}
 \centering
 \includegraphics[width = 0.4\textwidth]{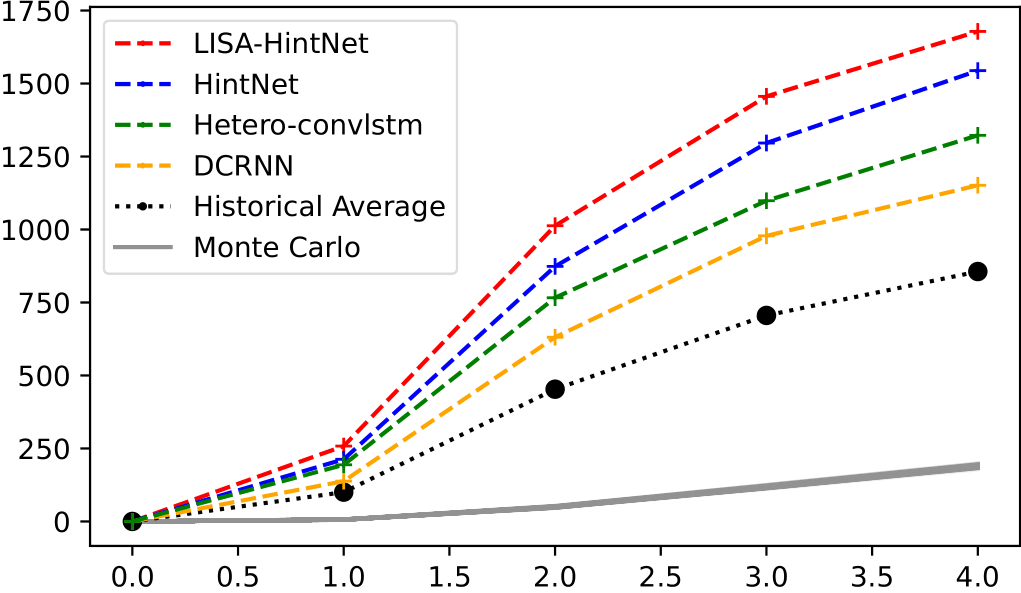}
 \caption{Cross-K function measuring spatial correlation between prediction and ground truth}
 \label{fig: cross}
\end{figure}

\subsection{Ablation study}
\label{ablation}

\subsubsection{Impact of Feature Groups}
To examine the contributions of feature groups on the model accuracy, we compare the performance of LISA trained with the network from HintNet on different feature groups in regions with different levels of spatial heterogeneity. 

\begin{table}[ht]
\fontsize{9}{11}\selectfont
\centering
\caption{Impact of Feature Groups}
\setlength{\tabcolsep}{3mm}{
\begin{tabular}{c|c|c|c}\toprule
\multirow{1}{*}{\tiny{$MSE\times 10^{-3}$}} &  \textbf{$F_S$} & \textbf{$F_S$+$F_T$} & \textbf{$F_S$+$F_T$+$F_{ST}$}  \\\midrule

		$16 \times 16$ & $24.21$ & $20.11$ & $17.39$  \\
		\hline
        $32 \times 32$ & $6.537$ &$5.921$ & $5.426$  \\
        \hline
		$64 \times 64$ & $3.164$ &$2.803$ & $2.292$ \\
		\hline
        $128 \times 64$ & $2.395$ & $2.168$ & $\textbf{1.971}$  \\
        \hline

\bottomrule
\end{tabular}}
\label{table: feature}
\end{table}

As Table \ref{table: feature} demonstrates, the models trained with only the spatial feature group ($F_S$) perform roughly the same as the historical average, which indicates that spatial features can only capture some geographical differences in accident risks but barely reveal more dynamic correlations. With the temporal features involved ($F_T$), the model on the region with size $16\times16$ makes considerable improvements. %and models on other regions make some reasonable improvements. 
Given the most homogeneous downtown area, it shows that calendar features contribute more to downtown areas, which indicates that accident patterns in those areas are more influenced by human activities. Interestingly, models trained with spatial-temporal features ($F_{ST}$) make substantial improvements on regions with size $32\times32$ and $64\times64$. This indicates that the spatial-temporal features including weather information and real-time traffic conditions contribute more to suburban areas which involve highway systems with large high-speed traffic volumes. Finally, the model performances in the largest area $128\times64$ are still improving by adding feature groups, but the scale is smaller as errors are averaged by a greater number of cells covered. 

\begin{table}[t]
\begin{threeparttable}[b]
\vspace{-0.0in}
\caption{Ablation Study}
\vspace{0mm}
\label{tab:performance}
\vskip 0.0in
\begin{center}
\begin{small}
\begin{sc}
\begin{tabular}{p{2.0cm}p{1.2cm}p{1.2cm}p{1.2cm}p{1.2cm}}
\toprule
\multirow{1}{*}{\thead{\textbf{\tiny{$\textbf{MSE}\times 10^{-3}$}}}} &

\footnotesize{\tiny{$128\times64$}} & \footnotesize{\tiny{$64\times64$}} & \tiny{\tiny{$32\times32$}} & \footnotesize{\tiny{$16\times16$}} \\
\midrule
LSTM   & 2.730 & 3.311 &  8.543 &  26.44  \\
LISA\tiny{-LSTM}$^*$   & 2.477 & \underline{2.911} &  8.419 &  \underline{25.22}  \\
LISA\tiny{-LSTM}   & \underline{2.474} & 2.913 &  \underline{8.414} &  \underline{25.22}  \\

\midrule
ConvLSTM   & 3.917 & 4.420 &  8.103 &  19.74  \\
LISA\tiny{-ConvLSTM}$^*$   & 2.231 & 2.710 &  6.974 &  19.25  \\
LISA\tiny{-ConvLSTM}   & \underline{2.229} & \underline{2.707} &  \underline{6.965} &  \underline{19.16}  \\

\midrule
DCRNN   & 2.542 & 2.917 &  7.823 &  17.78 \\
LISA\tiny{-DCRNN}$^*$  & \underline{2.120} & 2.511 &  6.126 &  17.59 \\
LISA\tiny{-DCRNN}  & 2.126 & \underline{2.486} &  \underline{6.079} &  \underline{17.43} \\

\midrule
HintNet   & 2.111 & 2.535 &  6.029 &  17.44 \\
LISA\tiny{-HintNet}$^*$   & 1.976 & 2.300 &  5.438 &  \textbf{17.41} \\
LISA\tiny{-HintNet}   & \textbf{1.971} & \textbf{2.288} &  \textbf{5.423} &  \textbf{17.41} \\

\hline

\bottomrule
\end{tabular}
\begin{tablenotes}
    \item $*$: Without Spatial Gradient Search
\end{tablenotes}
%\begin{tablenotes}
%    \item [*] \textbf{LISA}   
%\end{tablenotes}
\end{sc}
\end{small}
\end{center}
\label{tab: abla}
\end{threeparttable}
\vspace{-0.2in}
\end{table}

\subsubsection{Impact of Spatial Gradient Search}

We conducted an ablation study on the effectiveness of the proposed spatial gradient search algorithm, and the results are shown in Table. \ref{tab: abla}. The second rows of each block indicate the results of our model without spatial gradient search. In this case, we simply include all locations covered within the expanding. We also demonstrate the performance of baseline models in the first rows of each block. The results show that our proposed spatial gradient search algorithm can effectively improve the model performance in all regions with different levels of spatial heterogeneity.

\subsubsection{Impact of Parameters}
\begin{figure}[t]
    \centering
    
    \begin{minipage}{0.15\textwidth}
    \hspace{0cm}
    \includegraphics[width=1.0\textwidth]{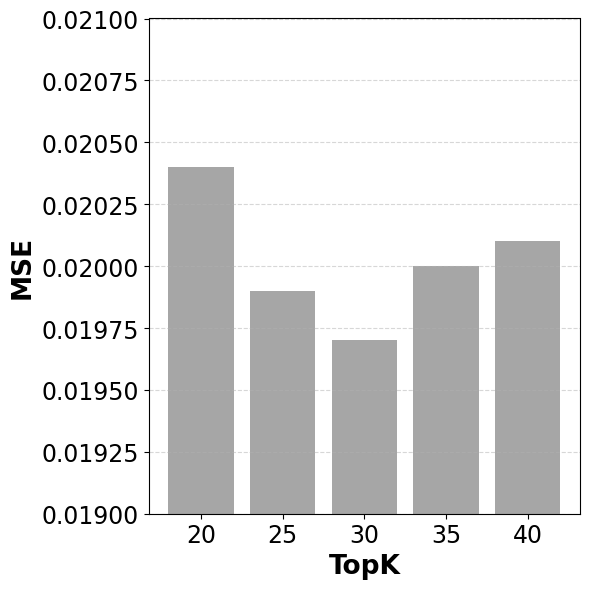}\hfill
    \caption*{\footnotesize {(a) TopK Locations}}
    \end{minipage}\hfill
    \begin{minipage}{0.15\textwidth}
    \includegraphics[width=1.0\textwidth]{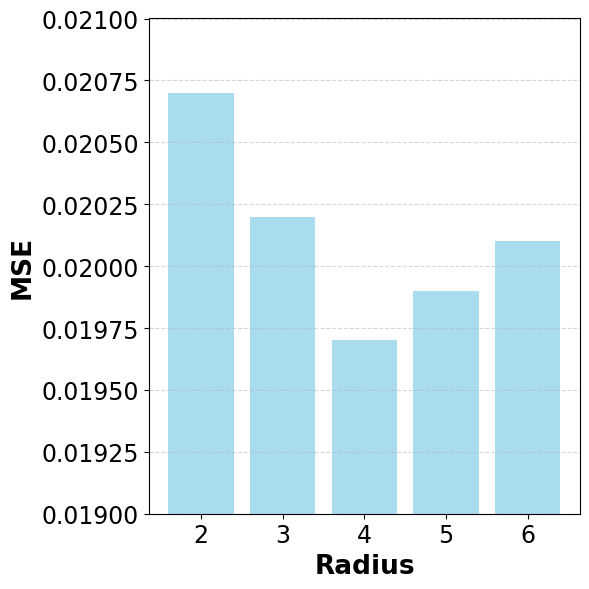}
    \caption*{\footnotesize{(b) Radius $r$}}
    \end{minipage}\hfill
    \begin{minipage}{0.15\textwidth}
    \includegraphics[width=1.0\textwidth]{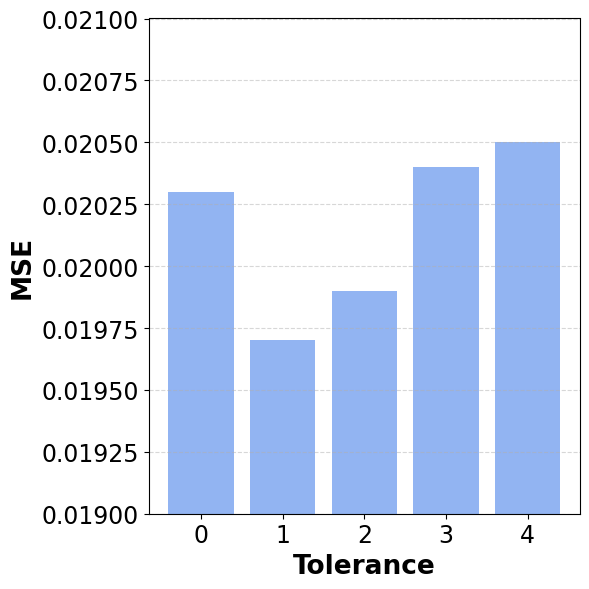}
    \caption*{\footnotesize{(c) Tolerance $\gamma$}}
    \end{minipage}\hfill
    
	\caption{\footnotesize{Performance on different parameter settings}}
	
	\label{three_data_map}
\end{figure}

We compare the performance of the proposed frameworks with different parameter settings, and the results are shown in Figure \ref{three_data_map}. First, we adjust the parameter, Topk, in expanding to adjust the size of locations selected in each expanding. This design avoids the problem of training a model on an extremely small candidate area. It arrives at the best performance when the Top-K equals $30$ as illustrated in Fig. \ref{three_data_map} (a). Second, we change the radius to control the expanding rate. With a smaller radius, the candidate areas tend to be smaller with fewer training samples, thus it becomes difficult to train the model. However, with a large radius, we sample the excessive number of candidates, so that it fails to capture the most appropriate partition. 
It reaches the best performance when the radius equals 4 in Fig. \ref{three_data_map} (b). Third, we also investigate the influence of different tolerance limits in Fig. \ref{three_data_map} (c). It starts from zero, which means no tolerance is allowed and it terminates sampling candidates immediately when error increases. In this case, it might fail to capture a potential better partition, therefore cannot reach optimum performance. 
It arrives at best results at tolerance one, and error rises with greater tolerance. This might be because the model is non-deterministic, thus excessive attempts eventually increase the probability of covering too many candidates.

\subsection{Case Study} 

\begin{figure}[ht]  \hspace*{0cm}
 \centering
 \includegraphics[width = 0.48\textwidth]{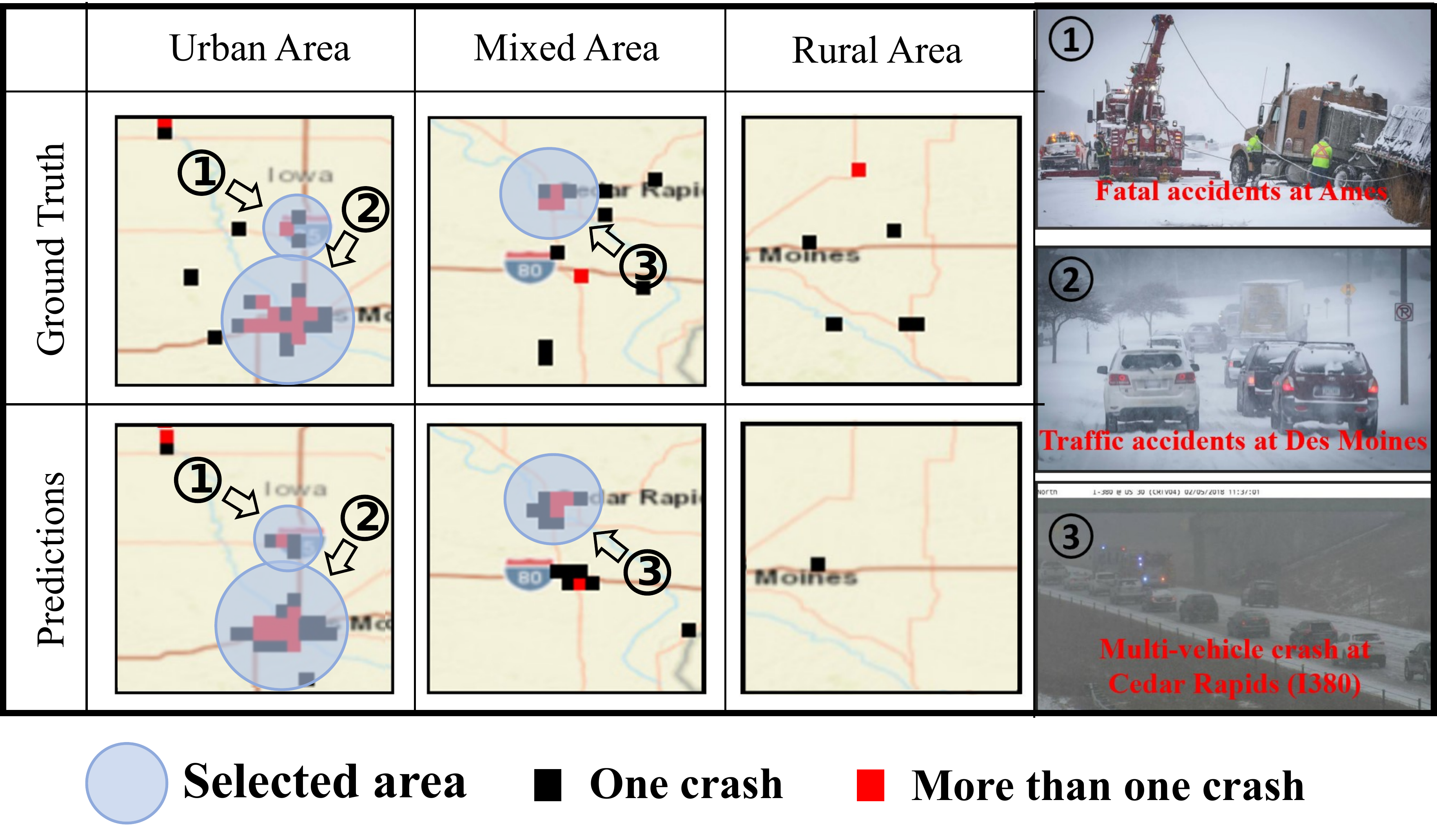}
 \caption{Case study on traffic accidents on Feb. $5^{th}, 2018.$}
 \label{fig: case}
\end{figure}

We show an example of a successful prediction by LISA on Feb $5^{th}$ 2018 in Figure \ref{fig: case}. There was a severe winter storm in the state of Iowa, and nearly 200 vehicles crashed on that date. Figure \ref{fig: case} shows the results in three selected regions from the urban area to the rural area. 

As observed, a majority of accidents and the number of crashes are correctly predicted in all regions. The blue circles highlight three major vehicle crash events that happened.  In the selected urban area, severe multi-vehicle crashes were reported in Des Moines and Ames along Interstate 35 \cite{2018des}\cite{2018cid} as shown on the right side of the first and second sub-figures. In addition to the locations of the traffic accidents, the red hotspots demonstrate that the frequency of the accidents in each grid cell is correctly predicted and matched with ground truth data.

\vspace{-0.1in}

\section{Conclusion}

In this paper, we tackled the traffic accident forecasting problem. Predicting future traffic accidents is important to traffic management and public safety. Accurate predictions could reduce the accident fatality rate and police response time. Related approaches mostly lack the ability to capture heterogeneous patterns across regions. Recent works solve this issue by using spatial ensembles, but it requires extra knowledge on accurately defining partitions. To address those limitations, we proposed a Learning-integrated Space Partition Framework that can work with any deep learning network and obtains partitioned sub-regions and trained models simultaneously. We performed comprehensive evaluations on a real-world dataset. The results demonstrated that our proposed work can automatically capture the inherent spatial heterogeneity and significantly improve the baseline model performance by an average of $13.0\%$. 

The designed framework is concentrated to traffic accident forecasting problem. In the future, we will investigate how to generalize our proposed mechanism to other applications and spatial problems.

\bibliographystyle{IEEEtran}
\bibliography{sample-base}

\end{document}